\documentclass{article}

\usepackage{spconf,amsmath,graphicx}
\usepackage{enumitem}
\usepackage{adjustbox}
\usepackage{threeparttable}
\usepackage{multirow}
\PassOptionsToPackage{hyphens}{url}\usepackage{hyperref}

\usepackage{xcolor,colortbl}
\definecolor{Gray}{gray}{0.85}
\newcolumntype{f}{>{\columncolor{Gray}}c}

\title{Automatic Detection of Sentimentality from Facial Expressions}
\name{Mina Bishay, Jay Turcot, Graham Page and Mohammad Mavadati}
\address{Smart Eye AB}
\begin{document}
%\ninept
%
\maketitle
%

%%%%%%%%%%%%%%%%%%%%%%%%%%%%%%%%%%%%%%%%%%%%%%%%%%%%%%%%%%%%%%%%%%%%%%%%%%%%%%%%
\begin{abstract}

Emotion recognition has received considerable attention from the Computer Vision community in the last 20 years. However, most of the research focused on analyzing the six basic emotions (e.g. joy, anger, surprise), with a limited work directed to other affective states. In this paper, we tackle sentimentality (strong feeling of heartwarming or nostalgia), a new emotional state that has few works in the literature, and no guideline defining its facial markers. To this end, we first collect a dataset of 4.9K videos of participants watching some sentimental and non-sentimental ads, and then we label the moments evoking sentimentality in the ads. Second, we use the ad-level labels and the facial Action Units (AUs) activation across different frames for defining some weak frame-level sentimentality labels. Third, we train a Multilayer Perceptron (MLP) using the AUs activation for sentimentality detection. Finally, we define two new ad-level metrics for evaluating our model performance. Quantitative and qualitative results show promising results for sentimentality detection. To the best of our knowledge this is the first work to address the problem of sentimentality detection. 

\end{abstract}

%%%%%%%%%%%%%%%%%%%%%%%%%%%%%%%%%%%%%%%%%%%%%%%%%%%%%%%%%%%%%%%%%%%%%%%%%%%%%%%%
%%%%%%%%%%%%%%%%%%%%%%%%%%%%%%%%%%%%%%%%%%%%%%%%%%%%%%%%%%%%%%%%%%%%%%%%%%%%%%%%

%%%%%%%%%%%%%%%%%%%%%%%%%%%%%%%%%%%%%%%%%%%%%%%%%%%%%%%%%%%%%%%%%%%%%%%%%%%%%%%%
%%%%%%%%%%%%%%%%%%%%%%%%%%%%%%%%%%%%%%%%%%%%%%%%%%%%%%%%%%%%%%%%%%%%%%%%%%%%%%%%
\section{INTRODUCTION}
%--------------------------------------------------------------------------------

Understanding facial expressions are quite important to analyze humans' emotions and non-verbal communications. Automatic Facial Expression Analysis (AFEA) has been an active research area in Computer Vision, as it has gained popularity in several applications like ad testing \cite{mcduff2014predicting, mcduff2016applications, efremova2020understanding}, driver state monitoring \cite{mualuaescu2019improving, wilhelm2019towards}, and health care \cite{jaiswal2017automatic, bishay2019schinet, bishay2019can}). In ad testing, analyzing customers' facial responses gives traders insights about customers engagement, liking, and purchase intent \cite{mcduff2014predicting}. However, AFEA is quite limited to the detection of AUs and basic emotions, as collecting and labelling real-world data for other emotional states are quite challenging.

%  in the last two decades

% The work done beyond the basic emotions is limited,
% mcduff2020some, mcduff2017new
% dua2019autorate
% syed2018automated, girard2014nonverbal
%--------------------------------------------------------------------------------

% The 6 basic emotions (happiness, sadness, fear, angry, surprise, disgust) are connected and can be understood from facial expressions. EM-FACS \cite{} connects those emotions to combinations of facial muscle movements (aka AUs). However, the work done beyond the basic emotions is limited, as collecting and labelling real-world data for other affective states is quite challenging. In this work we investigate a new affective state (i.e. sentimentality), and have a collected a real-world dataset using a web-based framework that records participants while watching commercial ads. 

%--------------------------------------------------------------------------------

Evoking emotions like sentimentality (an emotion with heartwarming or nostalgic feelings) in commercial ads is an emerging trend in advertising \cite{kaliouby}. Subsequently, there has been a growing interest in studying sentimentality \cite{mcduff2016discovering, mcduff2017large}. In \cite{mcduff2016discovering}, McDuff highlighted the prominent AUs in sentimental responses, while in \cite{mcduff2017large} McDuff classified the ad media content into 5 classes (informed, inspired, sentimental, amused, persuaded) based on the participants facial responses and some media features. The facial markers of sentimentality has not been defined in the literature, in contrast to the basic emotions that were interpreted from AUs through the Emotional Facial Action Coding System (EMFACS) \cite{friesen1983emfacs}. Therefore, it was difficult to directly label and detect sentimentality.

\begin{figure*} [t!]
  \centering{\includegraphics[width=0.999 \linewidth]{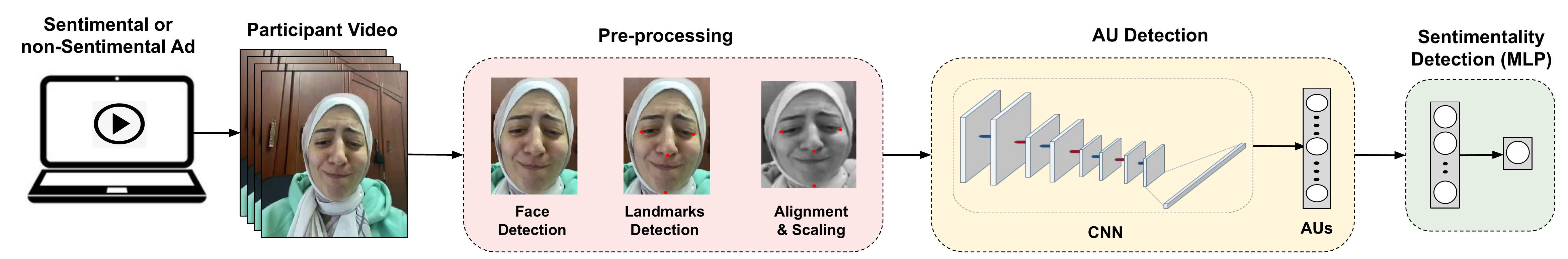}}
  \caption{The proposed architecture for sentimentality detection.}
  \label{fig_net}
\end{figure*}

In this paper, we present a novel methodology for detecting sentimentality. Specifically, we first collect real-world facial responses for participants watching a group of sentimental and non-sentimental ads. Second, we label the moments evoking sentimentality in the ads. Third, using the ad-level labels and an AU detector (predicting 20 different AUs), we filter and categorize the participants' frames in the training set into positive and negative examples. Specifically, frames with active AUs shown during sentimental moments are considered positive sentimentality examples, while other frames negative examples. Finally, we use the frame-level labels and AU predictions for training a MLP, that extracts high-level features on the top of the AU predictions (low-level features) for sentimentality detection. Fig. \ref{fig_net} shows an overview of our architecture. We believe that the proposed methodology can be replicated to other untackled emotions without the need for the exhaustive frame-level labelling, nor predefined facial markers for the target emotion.

%--------------------------------------------------------------------------------
%%%%%%%%%%%%%%%%%%%%%%%%%%%%%%%%%%%%%%%%%%%%%%%%%%%%%%%%%%%%%%%%%%%%%%%%%%%%%%%%
%%%%%%%%%%%%%%%%%%%%%%%%%%%%%%%%%%%%%%%%%%%%%%%%%%%%%%%%%%%%%%%%%%%%%%%%%%%%%%%%
\begin{figure*} [t!]
  \centering{\includegraphics[width=0.9 \linewidth]{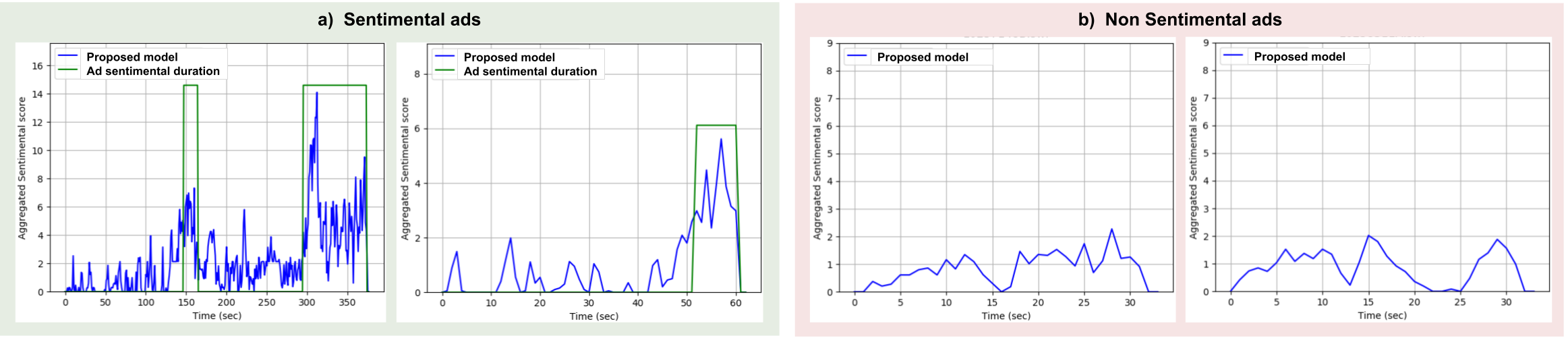}}
%   \centering{\includegraphics[width=0.99 \linewidth]{Paper_aggregate_curves_1.png}}
  \caption{The aggregated sentimentality across different sentimental and non-sentimental ads.}
  \label{fig_agg}
\end{figure*}

%--------------------------------------------------------------------------------
%%%%%%%%%%%%%%%%%%%%%%%%%%%%%%%%%%%%%%%%%%%%%%%%%%%%%%%%%%%%%%%%%%%%%%%%%%%%%%%%
%%%%%%%%%%%%%%%%%%%%%%%%%%%%%%%%%%%%%%%%%%%%%%%%%%%%%%%%%%%%%%%%%%%%%%%%%%%%%%%%

For evaluating our architecture, we define two new ad-level Key Performance Indicators (KPIs), that are based on the ad-level aggregated sentimentality. The first KPI measures how separable are the sentimental and non-sentimental ads in terms of the aggregated sentimentality, while the second measures if the aggregated sentimentality is firing high at the right sentimental moments. Our architecture shows promising qualitative and quantitative results for sentimentality detection. We base our emotion/sentimentality analysis on the recommendations given by Barrett {\em et al.} in \cite{barrett2019emotional} for studying facial movements in real life, sampling across different cultures, and using multiple facial stimuli. 

% We believe that the proposed methodology can be replicated to other untackled emotions without the need for the exhaustive frame-level labelling, nor pre-defined facial markers for the target emotion.

% Following the recommendations given by Barrett {\em et al.} in \cite{barrett2019emotional}, we base our analysis on   for studying facial movements in real life, sampling across different cultures, and using multiple facial stimuli. 

% compared to the chance level and any single AU
% Our experimental results also highlights some combinations of AUs that relate to sentimentality, and are aligned with the definition of sentimentality. 
%  across different ads
%--------------------------------------------------------------------------------
To the best of our knowledge this is the first work to directly address the problem of sentimentality detection. It is worth noting that in this paper we detect ``sentimentality'', a kind of emotion with heartwarming or nostalgic feelings, from an image capturing participant’s facial expression -- this is completely different from ``sentiment analysis'' that has a quite good amount of work in the literature \cite{medhat2014sentiment, zhang2018deep}, and basically detects if a piece of writing (e.g. reviews, survey responses) was positive, negative, or neutral.

The rest of the paper is organized as follows: In Section 2 and Section 3, we present the sentimentality dataset and the ad-level KPIs used in our analysis, respectively. In Section 4, we introduce the proposed methodology for sentimentality detection. Finally, we draw our conclusions in Section 5.

%--------------------------------------------------------------------------------
%%%%%%%%%%%%%%%%%%%%%%%%%%%%%%%%%%%%%%%%%%%%%%%%%%%%%%%%%%%%%%%%%%%%%%%%%%%%%%%%
%%%%%%%%%%%%%%%%%%%%%%%%%%%%%%%%%%%%%%%%%%%%%%%%%%%%%%%%%%%%%%%%%%%%%%%%%%%%%%%%
%%%%%%%%%%%%%%%%%%%%%%%%%%%%%%%%%%%%%%%%%%%%%%%%%%%%%%%%%%%%%%%%%%%%%%%%%%%%%%%%
%%%%%%%%%%%%%%%%%%%%%%%%%%%%%%%%%%%%%%%%%%%%%%%%%%%%%%%%%%%%%%%%%%%%%%%%%%%%%%%%

\section{Sentimentality Dataset}
%--------------------------------------------------------------------------------

Affectiva/SmartEye in collaboration with global market agencies have collected and analyzed thousands of commercial ads across different markets. For each ad, participants were hired to watch the ad, and then fill a survey about how they feel about the ad. A consent was given by the participants to get video recorded while they were watching the ad. The participants' facial responses in the videos were detected and analyzed to get insights about their level of engagement, liking, and purchase intent \cite{mcduff2014predicting}. 
% \textcolor{red}{Note that the participants were only informed about watching an ad for market research purposes, without defining the ad main driving emotion.}
% Note that the participants were not informed about the emotion the ad is trying to evoke.

% those responses are
% participants emotions and expressiveness.  Analyzing ads (i.e. ad testing) helps merchants get insights
%--------------------------------------------------------------------------------

For our analysis, experienced ad testers have selected 33 ads (18 sentimental and 15 non-sentimental), and 4.9K participants' videos to form a dataset for sentimentality. Sentimental ads are evoking sentimentality at some moments, while non-sentimental ads are typically informative, funny, or musical ads. The selected sentimental and non-sentimental ads span different markets (USA, UK, East and South Asia and Latin America), and subsequently the participants' videos used in our analysis have diverse demographics (gender, age band and ethnicity). Note that different participants were recruited for watching the different ads, and those participants were not informed about the emotion the ad is trying to evoke. The start and the end of the sentimental moments in the sentimental ads were labelled by 3 labellers, who were labelling by just watching the ads.

% Out of the thousands of ads been collected, e
% In order to form a dataset for sentimentality, we have selected 18 sentimental and 15 non-sentimental ads from Affectiva's large dataset.
% (the start and the end of the sentimental durations)
%--------------------------------------------------------------------------------

% non-sentimental ads not evoking sentimentality in their content. 
% The non-sentimental ads are divided into 20 non-funny ads and 10 funny ads. The reason we included funny ads is that funny-related expressions (e.g. smile) overlap with the sentimental expressions, and we want to test how the developed architecture (i.e. discovered expressions) can differentiate between the two categories. 

%--------------------------------------------------------------------------------

We use 3 sentimental ads for training our model, and 15 sentimental and 15 non-sentimental ads for testing. More ads/samples are needed in the testing as the proposed KPIs are calculated on the top of the aggregated sentimentality across different ads. For the participants' videos, we have around 2.1K videos for the 18 sentimental ads (250 videos for the training ads and 1.85K videos for the testing ads), and 2.8K videos for the 15 non-sentimental ads.
% -- these numbers after discarding videos with low face coverage

% Figure 1 shows the distribution of the ads across the different markets.

%--------------------------------------------------------------------------------

% Our analysis is based on using commercial ads that have been watched by a number of participants. 
% Our dataset consists of 50 commercial ads divided into 20 sentimental ads, 30 non-sentimental ads. Sentimental ads are those reflecting sentimentality at some moments, while non-sentimental ads are ads with no sentimental moments in their content, they are typically informative, funny, or musical ads. The non-sentimental ads are divided into 20 non-funny ads and 10 funny ads. The reason we included funny ads is that funny-related expressions (e.g. smile) overlap with the sentimental expressions, and we want to test how the developed architecture (i.e. discovered expressions) can differentiate between the two categories. Around 5K participants have watched the 50 commercial ads. Those participants were recorded using a web-based framework. The 5K face videos are filtered to include only videos with face tracking \geq90\% 

%--------------------------------------------------------------------------------
%%%%%%%%%%%%%%%%%%%%%%%%%%%%%%%%%%%%%%%%%%%%%%%%%%%%%%%%%%%%%%%%%%%%%%%%%%%%%%%%
%%%%%%%%%%%%%%%%%%%%%%%%%%%%%%%%%%%%%%%%%%%%%%%%%%%%%%%%%%%%%%%%%%%%%%%%%%%%%%%%

\section{Ad-level KPIs}
%--------------------------------------------------------------------------------

% In order to evaluate our architecture, we generate aggregated sentimentality curve for each ad, that is, a single curve produced by averaging the participants predicted sentimentality for each ad (as shown in Figure 1). The aggregate curves across sentimental and non-sentimental ads are used for calculating new ad-level KPIs. Specifically, we first use the ROC-AUC for measuring how separable are the maximum values across aggregate curves of the sentimental and non-sentimental ads. In addition, we test if the aggregated sentimentality score is firing high at the right sentimental moments. 

Our dataset does not have frame-level ground truth labels for sentimentality, so it is challenging to use the typical frame-level KPIs in our analysis. As the dataset has mainly ad-level labels defining the sentimental and non-sentimental ads, and the sentimental moments in the sentimental ads, we define new KPIs based on the available ad-level labels. Specifically, we aggregate (i.e. average) the participants' predicted sentimentality across each sentimental and non-sentimental ad, in order to get a single sentimentality curve for each ad (Fig. \ref{fig_agg} shows the aggregated sentimentality across four ads). Then, we calculate two KPIs that compare the ad-level predicted sentimentality to the ad-level labels.  

% Examples for aggregated sentimentality across four ads are given in Figure \ref{fig_agg}. 
% In order to have an ad-level sentimentality prediction that we can compare to the ad-level labels, we aggregate (i.e. average) the participants predicted sentimentality across each sentimental and non-sentimental ad, so as to get a single sentimentality curve for each ad. Examples for aggregated sentimentality across two ads are given in Figure \ref{fig_agg}.
% across different the participants watching an ad to form an ad-level sentimentality prediction.   

%--------------------------------------------------------------------------------

The first KPI, named \textbf{\textit{ROC-Ad}}, uses the area under the ROC curve (ROC-AUC) for measuring how separable are the sentimental and non-sentimental ads. To do so, we first calculate the maximum sentimentality score across the aggregate curve of each ad -- this leads to 15 scores across sentimental ads (considered the positive predictions), and 15 scores across non-sentimental ones (the negative predictions). Then, the ROC-AUC is calculated between the positive and negative predictions. 

% ROC-Ad indicates how separable are the sentimental and non-sentimental ads. 
% to have high activation for sentimental ads (ideal close to 1), and low activation for non-sentimental ads (ideal close to 0) -- this goal is measured by using a new KPI based on ROC-AUC.
% Having the ad-level labels and aggregated sentimentality curves, 
% The facial responses of all the participants are aggregated to form a single curve (called in our analysis "aggregate curve") for each ad. 
%--------------------------------------------------------------------------------

The second KPI, named \textbf{\textit{ROC-Sent}}, measures if the model is firing high at the right sentimental moments. Specifically, ROC-Sent uses ROC-AUC for measuring how separable are the sentimental and non-sentimental moments in the sentimental ads. Similar to the ROC-Ad, we calculate the maximum sentimentality score across the 15 sentimental moments (positive predictions), and the 15 non-sentimental moments (negative predictions). Then, the ROC-AUC is calculated between the positive and negative predictions.

\begin{table*}[ht]
\caption{Comparing the ROC-AUC of the developed AU detector to AFFDEX-SDK \cite{mcduff2016affdex}.}
% \caption{Comparing the proposed AU detection architecture to AFFDEX-SDK \cite{mcduff2016affdex}.}
\label{table_AUs}
\begin{adjustbox}{width=0.99 \textwidth, center}
  \centering
    \begin{tabular}{|c||c|c|c|c|c|c|c|c|c|c|c|c|c|c|c|c|c|c|c|c||f|}
 \hline
Facial & AU1 & AU2 & AU4 & AU5 & AU6 & AU7 & AU9 & AU10 & AU14 & AU15 & AU17 & AU18 & AU20 & AU24 & AU25 & AU26 & AU28 & Eye & Smile & Smirk & \textbf{Avg} \\ 
Expressions &  &  &  &  &  &  &  &  &  &  &  &  &  &  &  &  &  & closure & & &   \\ \hline 
% AFFDEX-SDK \cite{mcduff2016affdex} & 0.807 & 0.814 & 0.876 & \textbf{0.911} & 0.833 & 0.798 & 0.922 & 0.883 & \textbf{0.903} & 0.845 & 0.829 & 0.906 & 0.855 & 0.696 & 0.842 & 0.686 & 0.901 & \textbf{0.913} & 0.957 & 0.801 & 0.846 \\ \hline
AFFDEX \cite{mcduff2016affdex} & 0.76 & 0.79 & 0.86 & \textbf{0.87} & 0.92 & 0.75 & \textbf{0.91} & 0.86 & \textbf{0.86} & 0.78 & 0.79 & 0.91 & 0.86 & 0.76 & 0.86 & 0.63 & 0.91 & \textbf{0.92} & 0.94 & 0.82 & 0.84 \\ \hline
% \multirow{2}{*}{AFFDEX SDK \cite{mcduff2016affdex}} & AUC & 0.807 & 0.814 & 0.876 & \textbf{0.911} & 0.833 & 0.798 & 0.922 & 0.883 & \textbf{0.903} & 0.845 & 0.829 & 0.906 & 0.855 & 0.696 & 0.842 & 0.686 & 0.901 & \textbf{0.913} & 0.957 & 0.801 \\ \cline{2-22} 
%  & F1 & 0.338 & 0.0 & 0.272 & \textbf{0.288} & 0.001 & 0.032 & 0.045 & 0.068 & 0.002 & 0.0 & 0.075 & 0.015 & 0.0 & 0.0 & 0.005 & 0.0 & 0.349 & 0.001 & 0.607 & \textbf{0.111} \\ \hline  \hline

% Proposed method & \textbf{0.865} & \textbf{0.909} & \textbf{0.940} & 0.897 & \textbf{0.980} & \textbf{0.944} & \textbf{0.958} & \textbf{0.950} & 0.836 & \textbf{0.875} & \textbf{0.864} & \textbf{0.950} & \textbf{0.928} & \textbf{0.875} & \textbf{0.904} & \textbf{0.811} & \textbf{0.963} & \textbf{0.913} & \textbf{0.975} & \textbf{0.866} & \textbf{0.910}  \\ \hline
Ours & \textbf{0.79} & \textbf{0.84} & \textbf{0.92} & 0.85 & \textbf{0.94} & \textbf{0.83} & 0.89 & \textbf{0.93} & 0.80 & \textbf{0.88} & \textbf{0.88} & \textbf{0.92} & \textbf{0.93} & \textbf{0.87} & \textbf{0.89} & \textbf{0.71} & \textbf{0.96} & 0.91 & \textbf{0.97} & \textbf{0.84} & \textbf{0.88}  \\ \hline
% \multirow{2}{*}{Proposed method} & AUC & \textbf{0.865} & \textbf{0.909} & \textbf{0.940} & 0.897 & \textbf{0.980} & \textbf{0.944} & \textbf{0.958} & \textbf{0.950} & 0.836 & \textbf{0.875} & \textbf{0.864} & \textbf{0.950} & \textbf{0.928} & \textbf{0.875} & \textbf{0.904} & \textbf{0.811} & \textbf{0.963} & \textbf{0.913} & \textbf{0.975} & \textbf{0.866}  \\ \cline{2-22}
%  & F1 & \textbf{0.362} & \textbf{0.365} & \textbf{0.576} & 0.222 & \textbf{0.581} & \textbf{0.156} & \textbf{0.092} & \textbf{0.156} & \textbf{0.187} & \textbf{0.109} & \textbf{0.312} & \textbf{0.112} & \textbf{0.153} & \textbf{0.120} & \textbf{0.580} & \textbf{0.257} & \textbf{0.404} & \textbf{0.518} & \textbf{0.673} & 0.085  \\ \hline

 \end{tabular}
\end{adjustbox}
\vspace{-4mm}
\end{table*}

\section{Method}
%--------------------------------------------------------------------------------

% labelling refinement part
In this section we present the proposed methodology for detecting sentimentality. Our model takes as input a frame depicting a face and gives as output a binary label indicating if the face is showing markers of sentimentality or not. The analysis is performed in 3 stages; preprocessing, AU detection, and sentimentality detection. Fig. \ref{fig_net} shows an overview of the whole architecture.

After collecting a dataset of diverse participants watching some sentimental and non-sentimental ads, and labeling the moments evoking sentimentality. We first detect and align the participant face. Then, we build an AU detector for analyzing the participant facial expression. The AU predictions are used for defining some weak frame-level sentimentality labels, as well as extracting low-level features from the face image. Finally, we train a MLP using the AU predictions for extracting high-level features for sentimentality detection.

\subsection{Preprocessing}
% Low-Level Feature Extraction: 
%--------------------------------------------------------------------------------

In order to process each video in our dataset, we first extract the  region  of  interest  (i.e. the  participant  face) at each frame, by using a face detector trained in the wild. To ensure we are including participants who are not distracted from the ad or away from the screen, we only include participants with face coverage (i.e. the percentage of the video frames with a face got detected) $\geq90\%$ for the next steps. Second, we extract 4 facial landmarks (outer eye corners, nose tip and chin) from the face region. These landmarks are used for aligning the face horizontally to have a zero roll angle. Finally, the aligned faces are scaled to a fixed resolution, and passed as an input to the AU detection architecture.

% aligning the face image, by rotating the line connecting the 2 eye corners to have a zero roll angle.
%  -- these numbers after discarding videos with low face coverage
% (i.e. reducing the head roll rotation)

%--------------------------------------------------------------------------------
%--------------------------------------------------------------------------------
%--------------------------------------------------------------------------------

\subsection{AU Detection}
% Low-Level Feature Extraction: 
%--------------------------------------------------------------------------------

% In this section we will explain describe the large-scale dataset used for training and testing the architecture, and then we describe the CNN, and experimental settings used in our analysis.

In this section we describe the dataset, CNN architecture, and experimental settings used for building and training our AU detector. Most of the settings in our AU detection architecture are chosen based on the recommendations given in \cite{bishay2021choose, bishay2021cnns}.

%--------------------------------------------------------------------------------

\textbf{AU Dataset.} In the literature, several datasets (e.g. DISFA \cite{mavadati2013disfa}, UNBC \cite{lucey2011painful}) have been used for training different architectures. However, many of these datasets have relatively limited number of participants, recording conditions, and/or diversity in demographics. In our analysis, we use a large-scale dataset consisting of $\sim55K$ videos, that were captured in the wild. The participants in our dataset have diverse age, gender and ethnicity. For the experiments, we divide the dataset into 40.9K videos for training, 5.9K for validation, and 8.2K for testing.

% , BP4D \cite{zhang2014bp4d}
%  (at different head poses and illumination conditions)

% DISFA+ \cite{mavadati2016extended}, 
% Spontaneous facial behavior, in contrast to posed or deliberate facial expressions, is generally more challenging to get annotated because of different timing and dynamics of AUs \cite{schmidt2006role}. 
% The size and quality of data matter. 
% In the past decade, several datasets have been collected and annotated for spontaneous behavior, e.g. DISFA \cite{mavadati2013disfa}, UNBC \cite{lucey2011painful} and BP4D \cite{zhang2014bp4d}. However, these datasets have a relatively limited number of participants, mainly controlled recording conditions, and/or diversity in demographics. For our analysis, we collected and labeled a large-scale face video dataset, that was captured in naturalistic conditions and with totally spontaneous facial expressions. This dataset contains $\sim$53 thousands (53K) face videos of participants with diverse age, gender and ethnicity distributions. For the experiments, we divided it into 40K videos for training, 5K videos for validation, and 8K videos for testing. 

% (in various lighting conditions and head poses)
%  and with totally spontaneous facial expressions
%--------------------------------------------------------------------------------

Our large-scale dataset was collected using the web-based approach described in \cite{mcduff2013affectiva, mcduff2018fed+}. The videos were collected worldwide (from 90+ countries) for participants watching commercial ads. The videos were manually annotated for the presence of 20 AUs using trained FACS coders. A part of this dataset was made available to the research community through AM-FED \cite{mcduff2013affectiva} and AM-FED+ \cite{mcduff2018fed+}. Note that the videos in the AU dataset are different from the ones in the sentimentality dataset.

% Moreover, the videos were labelled for gender (55\% Female, 37\% Male, 8\% uncertain) and ethnicity (37\% Caucasian, 24\% East Asian, 14\% South Asian, 13\% Latin, 9\% African, 3\% uncertain).
%--------------------------------------------------------------------------------

% , with each frame labeled by three coders
% Fig \ref{} shows example images for our large-scale dataset that is captured in diverse lighting conditions (i.e. in the wild).
% so as to have a rich dataset of facial expressions. 
%--------------------------------------------------------------------------------
%--------------------------------------------------------------------------------
%--------------------------------------------------------------------------------
%--------------------------------------------------------------------------------

\textbf{CNN architecture.} We treat the AU detection problem as a multi-label classification problem where a single CNN is jointly trained for detecting 20 AUs simultaneously (AUs are given in Table \ref{table_AUs}). The CNN consists of 5 convolutional and 1 fully-connected layers. A max-pooling layer is used after each convolutional layer. The fully-connected layer has 20 sigmoid units representing the predictions of the 20 AUs. 

% The 5 convolutional layers have 32, 64, 96, 128 and 256 filters respectively. All the convolutional filters have a kernel of size 3$\times$3.
% A max-pooling layer with a filter of size 2$\times$2 is used after each convolutional layer.
%--------------------------------------------------------------------------------
%--------------------------------------------------------------------------------
%--------------------------------------------------------------------------------
%--------------------------------------------------------------------------------

\textbf{Experimental settings.} As we are using a naturalistic dataset, most of the AUs in our dataset are severely imbalanced, having a high ratio of negative to positive examples. In order to avoid biasing the classifier to the most frequent class, we use oversampling to balance the data. The training batches are augmented by flipping, shifting, rotation, etc. The Binary Cross Entropy function is used for calculating the loss.

%  Specifically, we sample 4 positive and 4 negative frames for each AU, which results in 160 images for each training batch (20 AUs $\times$ 8 images). The positive and negative examples are chosen randomly from the training set.
% During training, the loss over each AU is calculated first, and then the total loss is calculated as the average of the different AU losses. 

% the loss over every AU is calculated, and then the total loss is calculated as the average of the different AU losses. 
% Note that using oversampling with a multilabel classifier retains the AUs equally represented and balanced in the training batches, however, it discards the correlations incorporated between the different AUs. 
% (the ratio of correctly classified images over the total number of images in each batch)

%--------------------------------------------------------------------------------

% The CNN is trained for 1200 epochs, and for each epoch 300 batches are sampled for training and 300 batches for validation. The training batches are augmented by flipping, shifting, rotation, etc. 
We compare the developed AU detector to a widely-used facial analysis toolkit (AFFDEX-SDK \cite{mcduff2016affdex}) on our large testing set. ROC-AUC is used for evaluating the AU detection performance. Table \ref{table_AUs} shows the performance across the different AUs for both models. Our AU detector achieves better performance than AFFDEX-SDK for most of the AUs (by $\sim$4\% on average).

\begin{table*}[ht]
\caption{Sentimentality evaluation results across the different AUs and the proposed model.}
\label{table_1}
\begin{adjustbox}{width=0.99 \textwidth, center}
  \centering
    \begin{tabular}{|c||c|c|c|c|c|c|c|c|c|c|c|c|c|c|c|c|c|c|c|c|c||f|}
 \hline

% KPIs & Chance & AU1 & AU2 & AU4 & AU5 & AU6 & AU7 & AU9 & AU10 & AU14 & AU15 & AU17 & AU24 & AU25 & AU26 & AU28 & Eye & Smile & Smirk   \\ 
%  & level &  &  &  &  &  &  &  &  &  &  &  &  &  &  &  &  Closure &  &    \\ \hline 

% ROC-Ad & 50 & 69 & 79 & 61 & 73 & 68 & 63 & 61 & 65 & 70 & 67 & 76 & 75 & 65 & 70 & 70 & 54 & 64 & 62  \\ \hline 
% ROC-Sent & 50 & 51 & 59 & 44 & 45 & 57 & 58 & 58 & 50 & 46 & 45 & 47 & 52 & 57 & 57 & 46 & 33 & 57 & 49   \\ \hline 
% ROC-Fun & 50 & 73 & 67 & 74 & 69 & 41 & 49 & 55 & 62 & 67 & 70 & 79 & 75 & 51 & 67 & 70 & 45 & 37 & 68   \\ \hline 

\textbf{KPIs} & Chance & AU1 & AU2 & AU4 & AU5 & AU6 & AU7 & AU9 & AU10 & AU14 & AU15 & AU17 & AU18 & AU20 & AU24 & AU25 & AU26 & AU28 & Eye & Smile & Smirk & \textbf{Proposed}  \\
& level &  &  &  &  &  &  &  &  &  &  &  &  &  &  &  &  & & Closure  & & &  \\ \hline

% Sent mean & 0.5 & 0.67 & 1.55 & 5.70 & 1.18 & 3.29 & 1.38 & 0.56 & 0.62 & 1.32 & 3.66 & 3.36 & 1.84 & 2.06 & 2.37 & 5.80 & 2.60 & 2.72 & 2.88 & 7.44 & 0.38 & 0.89 \\ \hline
% Non Sent mean & 0.5 & 0.46 & 1.23 & 4.94 & 1.00 & 2.08 & 1.42 & 0.63 & 0.65 & 1.14 & 2.69 & 2.52 & 1.59 & 1.30 & 2.01 & 5.51 & 1.97 & 2.31 & 3.20 & 5.21 & 0.32 & 0.68 \\ \hline \hline
% ROC Ad & 50 & 70 & 81 & 60 & 76 & 72 & 65 & 64 & 64 & 71 & 64 & 75 & 74 & 70 & 75 & 68 & 72 & 72 & 48 & 67 & 64 & 79  \\ \hline
% ROC Sent & 50 &  51 & 57 & 44 & 45 & 57 & 58 & 58 & 45 & 45 & 45 & 48 & 37 & 58 & 52 & 55 & 57 & 45 & 33 & 57 & 48 & 61 \\ \hline \hline
% Average & 50 & 60.5 & 69 & 52 & 60.5 & 64.5 & 61.5 & 61 & 54.5 & 58 & 54.5 & 61.5 & 55.5 & 64 & 63.5 & 61.5 & 64.5 & 58.5 & 40.5 & 62 & 56 & \textbf{70}   \\ \hline

ROC-Ad & 0.50 & 0.61 & 0.73 & 0.55 & 0.74 & \textbf{0.84} & 0.54 & 0.52 & 0.61 & 0.70 & 0.69 & 0.72 & 0.73 & 0.67 & 0.73 & 0.60 & 0.75 & 0.74 & 0.43 & 0.66 & 0.72 & 0.79 \\ \hline
ROC-Sent & 0.50 & 0.44 & 0.58 & 0.41 & 0.36 & 0.36 & 0.46 & 0.32 & 0.52 & 0.48 & 0.47 & 0.36 & 0.40 & 0.51 & 0.40 & 0.35 & 0.54 & 0.48 & 0.27 & 0.60 & 0.50 & \textbf{0.61} \\ \hline \hline
Avg & 0.50 & 0.52 & 0.65 & 0.48 & 0.55 & 0.60 & 0.50 & 0.42 & 0.56 & 0.59 & 0.58 & 0.54 & 0.56 & 0.59 & 0.56 & 0.47 & 0.64 & 0.61 & 0.35 & 0.63 & 0.61 & \textbf{0.70}   \\ \hline

% ROC Fun & 50 &  76 & 61 & 69 & 56 & 52 & 51 & 60 & 52 & 67 & 66 & 76 & 86 & 59 & 77 & 63 & 77 & 69 & 42 & 48 & 73  \\ \hline \hline
% Average & 50 & 65.67 & 66.33 & 57.67 & 59.00 & 60.33 & 58.00 & 60.67 & 53.67 & 61.00 & 58.33 & 66.33 & 65.67 & 62.33 & 68.00 & 62.00 & 68.67 & 62.00 & 41.00 & 57.33 & 61.67  \\ \hline

    \end{tabular}
\end{adjustbox}
\end{table*}

%--------------------------------------------------------------------------------
%%%%%%%%%%%%%%%%%%%%%%%%%%%%%%%%%%%%%%%%%%%%%%%%%%%%%%%%%%%%%%%%%%%%%%%%%%%%%%%%
%%%%%%%%%%%%%%%%%%%%%%%%%%%%%%%%%%%%%%%%%%%%%%%%%%%%%%%%%%%%%%%%%%%%%%%%%%%%%%%%

\begin{figure*} [t!]
  \centering{\includegraphics[width=0.82 \linewidth]{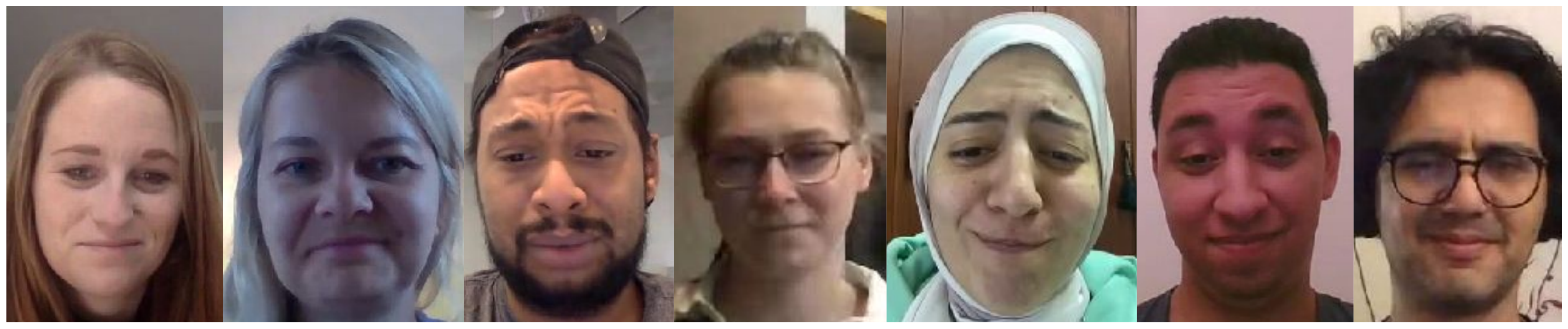}}
%   \centering{\includegraphics[width=0.99 \linewidth]{MLP_Sentimental_faces_first.pdf}}
  \caption{The detected positive moments of sentimentality using the proposed model.}
%   \caption{Sentimental faces detected by a MLP trained on expressions shown during sentimental moments.}
  \label{fig_1}
\end{figure*}

%--------------------------------------------------------------------------------
%%%%%%%%%%%%%%%%%%%%%%%%%%%%%%%%%%%%%%%%%%%%%%%%%%%%%%%%%%%%%%%%%%%%%%%%%%%%%%%%
%%%%%%%%%%%%%%%%%%%%%%%%%%%%%%%%%%%%%%%%%%%%%%%%%%%%%%%%%%%%%%%%%%%%%%%%%%%%%%%%

\subsection{Sentimentality detection}
%--------------------------------------------------------------------------------

We train a MLP on the top of the AU predictions for detecting sentimentality. For training the MLP, we need to have a set of positive and negative examples for sentimentality. There is no guideline in the literature that has defined the markers of sentimentality -- this is in contrast to basic emotions that are described by AUs through EMFACS \cite{friesen1983emfacs}. Subsequently, it is challenging to directly label a face dataset for sentimentality. In this paper we use the ad-level labels defining the sentimental moments for extracting positive and negative examples for sentimentality. Specifically, we use the expressive faces shown during sentimental moments as positive examples, and all the expressions shown during non-sentimental moments as negative examples. 

% extracting high-level features and
%--------------------------------------------------------------------------------

In order to make the positive examples cleaner, we first discard all the frames with no active AUs. Second, we evaluate using the ad-level KPIs how indicative are the 20 detected AUs for sentimentality (i.e. test if the activation of a single AU can be used as a marker for sentimentality). Table \ref{table_1} shows the results across the different AUs. Results show that most of the AUs are achieving relatively low performance on one or two of the KPIs. Subsequently, sentimentality is potentially expressed by more complex combination of AUs. Based on that, we discard all the positive frames with only one active AU. Eventually, positive examples include frames with $\geq$2 active AUs during sentimental moments, while the negative examples include any other frame (with or without active AUs) in the non-sentimental moments.

% a single AU, and includes
% , as well as for the random prediction of sentimentality
% on one or two of the ad-level KPIs
%--------------------------------------------------------------------------------

The MLP consists of 2 Fully-Connected (FC) layers, the first FC has 8 neurons for extracting high-level features on the top of the AU predictions, while the second has 1 neuron for detecting sentimentality. We train the MLP using the positive and negative examples extracted from the participants watching the 3 sentimental ads in the training set. We train the MLP for 100 epochs using the Adam optimizer. For testing, we use 15 sentimental and 15 non-sentimental ads. The detected sentimentality for different participants is aggregated for each ad in the testing set, and then the ad-level KPIs are calculated. 

% ($\sim$250 videos)
% with an initial learning rate set to 0.01
% The batch size is set to 256.
%--------------------------------------------------------------------------------

Table \ref{table_1} shows the ROC-Ad and ROC-Sent achieved by the proposed model. On average our model has better performance than the chance level and the 20 AUs. The MLP combines different AUs to get better representation for sentimentality. Fig. \ref{fig_agg} shows the aggregated sentimentality across 2 sentimental and 2 non-sentimental ads from the testing set, as well as the sentimental moments in the ads. The aggregate curves show that our model has relatively higher activation at the sentimental ads than the non-sentimental ones. In addition, the model is firing more accurately at the sentimental moments. Fig. \ref{fig_1} shows some of the faces with positive sentimentality detection (faces belong to Smart Eye employees who have been recorded while watching a sentimental ad). Reviewing the detected faces shows that sentimentality has different combinations of AUs, and is not expressed in the same way across different people. 

We believe that the proposed methodology can be replicated for other emotional states that has no predefined facial markers. In addition, collecting and labelling some stimuli (e.g. ads) evoking the target emotional state can replace the exhaustive frame-level labelling.

\section{Conclusion}

%--------------------------------------------------------------------------------

In this work we present a novel methodology for detecting sentimentality. Our architecture consists of 3 steps; a) detecting and aligning the participants' faces, b) detecting 20 different AUs for each frame (low-level features), and c) training a MLP for detecting sentimentality by extracting high-level features on the top of the AU predictions. For training our model, we first collect a dataset of participants watching some sentimental and non-sentimental ads, and then we label the moments evoking sentimentality in the ads. The ad-level labels along with the frame-level AU activations are used for defining a group of positive and negative examples for training the MLP. We define two new ad-level KPIs for evaluating our model performance, by measuring how separable are the sentimental and non-sentimental ads, and the sentimental and non-sentimental moments. Qualitative and quantitative results show promising results for sentimentality detection.

% The ad-level labels defining the sentimental moments in 3 sentimental ads are used for extracting a group of positive and negative examples for training the MLP for sentimentality detection.

% SentimeNet is tested on a group of 15 sentimental and 15 non-sentimental ads.
% The MLP uses ad-level labels defining the sentimental and non-sentimental moments in 3 sentimental ads for extracting a group of positive and negative examples for training the MLP for sentimentality detection.
% in terms of 20 AUs
% while the MLP is trained and tested using a group of sentimental and non-sentimental ads.
% using a group of sentimental and non-sentimental ads.  

%--------------------------------------------------------------------------------
%%%%%%%%%%%%%%%%%%%%%%%%%%%%%%%%%%%%%%%%%%%%%%%%%%%%%%%%%%%%%%%%%%%%%%%%%%%%%%%%
%--------------------------------------------------------------------------------

% \section{ACKNOWLEDGMENTS}
% %--------------------------------------------------------------------------------

% We would like to thank Matt Strafuss and Dr. Rana el Kaliouby for their continued support and technical guidance.

%%%%%%%%%%%%%%%%%%%%%%%%%%%%%%%%%%%%%%%%%%%%%%%%%%%%%%%%%%%%%%%%%%%%%%%%%%%%%%%%

\bibliographystyle{IEEEbib}
\bibliography{Template}

\end{document}